%% file: main.tex
\documentclass{article}
\usepackage{spconf,amsmath,graphicx,hyperref}

\usepackage{adjustbox}
\usepackage{framed}
\usepackage{rotating}
\usepackage{wrapfig}
\usepackage{makecell}
\usepackage{booktabs}       
\usepackage{multirow}
\usepackage{balance}

\usepackage{amssymb}
\usepackage{mathtools}
\usepackage{algorithmic}
\usepackage{algorithm}
\usepackage{amsthm}

\usepackage[table]{xcolor}
\usepackage{arydshln}
\definecolor{wrppeach}{HTML}{FCE8D8}

\newcommand{\wrpshade}{\cellcolor{wrppeach}}
\hypersetup{hidelinks}

\title{Forward-Free LLM Depth Pruning via Weight Redundancy}
\name{Vincent-Daniel Yun$^{1, \dag}$\sthanks{Corresponding author. \dag Equal Contribution}, Woosang Lim$^{2, \dag}$}
\address{$^1$University of Southern California, United States \\
$^2$Seoul National University, Republic of Korea
}
\begin{document}
%
\maketitle
\begin{abstract}
\input{contents/0-abstract}
\end{abstract}


\section{Introduction}
\input{contents/1-intro}

\section{Related work}
\input{contents/2-related}

\section{Forward-Free Weight Redundancy}
\input{contents/3-method}

\section{Experimental Results}
\input{contents/4-results}

\section{Conclusion}
\input{contents/6-conclusion}

\clearpage

\bibliographystyle{IEEEbib}
\begingroup
\interlinepenalty=10000
\bibliography{refs}
\endgroup

\end{document}

%% file: contents/0-abstract.tex
Depth pruning reduces large language model (LLM) inference cost by removing complete Transformer blocks. Activation-based methods collect hidden states through forward passes on calibration data, while existing forward-free methods score each Transformer block separately without measuring similarity between blocks. We propose Weight-Redundancy Pruning (WRP), a forward-free depth-pruning method that estimates inter-layer redundancy from checkpoint weights to select blocks without calibration data or model forward passes. WRP compares attention output and MLP down-projection weights across layers and combines their pairwise similarities with relative projection-scale information. The resulting all-pairs similarity matrix guides layer grouping and block selection. Across multiple pruning settings, model families, and downstream tasks, WRP consistently outperforms existing forward-free magnitude pruning and approaches the performance of activation-based methods.

%% file: contents/1-intro.tex
\label{sec:introduction}

As large language models (LLMs) continue to improve, their increasing size also raises the memory and computation required for inference. In practice, pretrained model families are typically released at only a limited set of fixed model sizes, which may not match the resource budget of a user's hardware. Depth pruning provides a flexible alternative by removing complete Transformer blocks from an existing checkpoint, allowing the model size to be adjusted to the available memory and computation budget while retaining standard dense operators~\cite{kim2024shortened,men2025shortgpt}.

However, many depth-pruning methods require running forward passes through the original model to select layers for removal. Activation-based methods use these passes to collect hidden states on calibration data. ShortGPT compares block input-output states~\cite{men2025shortgpt}, LLM-Streamline compares boundary activations~\cite{chen2025streamline}, and LoRP computes all-pairs activation similarities~\cite{yun2026lorp}. For large models, running calibration forward passes and storing activations can exceed the available GPU memory before any blocks are removed. Thus, a GPU with enough memory to run the pruned model may still be unable to support the pruning process on the original model.

Forward-free pruning avoids this requirement by selecting layers directly from the checkpoint without calibration forward passes. Existing forward-free criteria, such as Mag and Mag+ in Shortened LLaMA, rely on weight magnitudes and score each block independently~\cite{kim2024shortened}. However, magnitude reflects the scale of an individual block rather than redundancy between layers. This limitation distinguishes magnitude-based criteria from activation-based methods that explicitly measure inter-layer similarity.

\begin{figure}[H]
    \centering
    \includegraphics[width=\columnwidth]{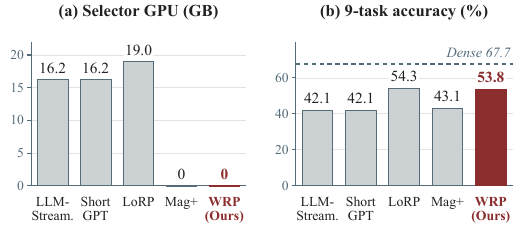}
    \caption{Selection GPU memory and nine-task accuracy on LLaMA-3.1-8B at 25\% depth pruning. Forward-free WRP retains accuracy close to LoRP without calibration forward passes.}
    \label{fig:intro}
\end{figure}

\begin{figure*}[t]
    \centering
    \includegraphics[width=\textwidth,trim=0 30bp 0 0,clip]{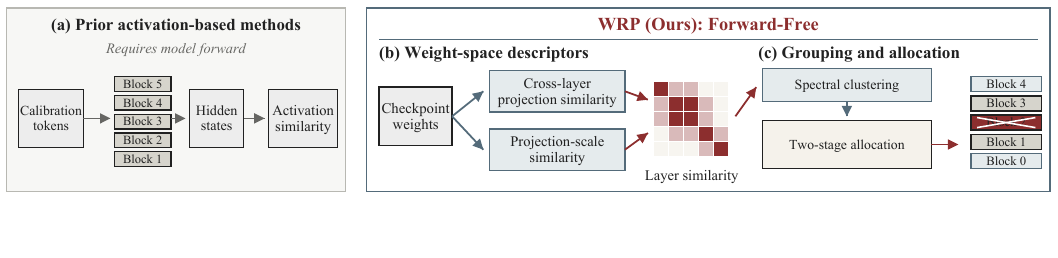}
    \caption{Forward-free WRP and an activation-based comparison. (a) The activation path is shown for comparison and is not part of WRP. (b)--(c) WRP constructs pairwise weight similarity and selects blocks globally without calibration data or model forward passes. Matrix entries and removed blocks are schematic.}
    \label{fig:overview}
\end{figure*}

Therefore, we propose Weight-Redundancy Pruning (WRP), a forward-free method that estimates pairwise inter-layer redundancy directly from checkpoint weights. WRP combines similarity between attention output and MLP down-projection weights with projection-scale information to construct an all-pairs similarity matrix for global layer selection. This enables depth pruning without calibration data, model forward passes, gradients, or internal-channel matching (Fig.~\ref{fig:overview}).

Across six pruning settings on three model families and nine tasks, WRP substantially outperforms forward-free magnitude pruning while remaining close to activation-based methods without post-pruning recovery. These results show that checkpoint weights alone provide useful inter-layer redundancy signals for effective depth pruning.

%% file: contents/2-related.tex
Pruning reduces model cost by removing parameters or larger structural components. Unstructured pruning removes individual weights, where magnitude-based criteria are among the most widely used approaches due to their simplicity and efficiency~\cite{frantar2023sparsegpt,sun2024wanda, 10650301}. Structured pruning instead removes larger model components~\cite{ma2023llmpruner,ashkboos2024slicegpt}. For block-level LLM depth pruning, Shortened LLaMA's Mag+ selects complete Transformer blocks directly from checkpoint weights without calibration forward passes~\cite{kim2024shortened}, but scores each block independently.

Activation-based depth-pruning methods use calibration forward passes to estimate layer redundancy. ShortGPT compares block input-output states~\cite{men2025shortgpt}, while LLM-Streamline compares boundary activations~\cite{chen2025streamline}. LoRP constructs global pairwise activation similarities for layer clustering and redundancy-based allocation~\cite{yun2026lorp}. These methods capture relations between layers but require model execution and calibration data. WRP instead estimates pairwise inter-layer redundancy directly from checkpoint weights without calibration forward passes.
\input{contents/4-results-full-table}

\begin{figure*}[t]
    \centering
    \includegraphics[width=0.98\textwidth]{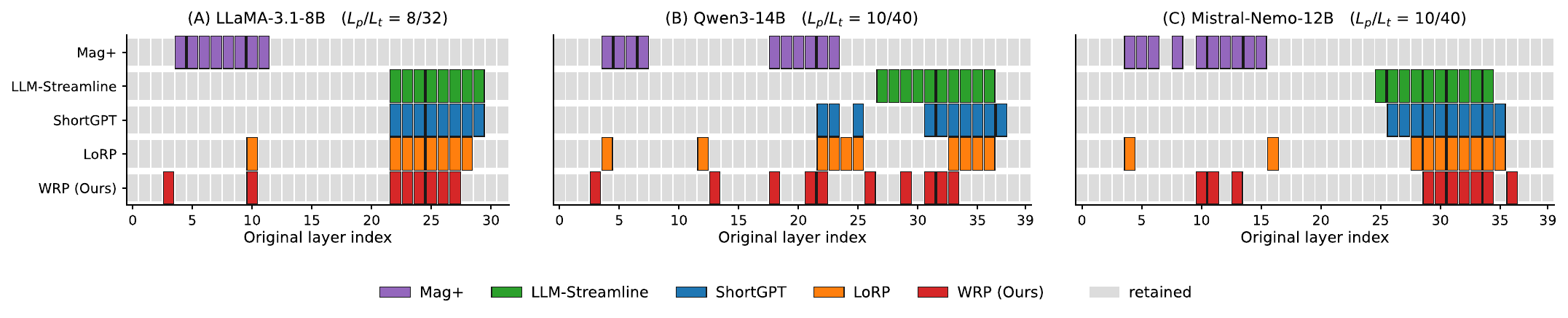}
    \caption{Layer-removal patterns at 25\% depth reduction: 8/32 blocks for LLaMA and 10/40 for Qwen and Mistral. Colored cells mark removed blocks; gray cells are retained. WRP selects blocks using checkpoint weights alone.}
    \label{fig:patterns}
    \par\medskip
    \includegraphics[width=0.98\textwidth]{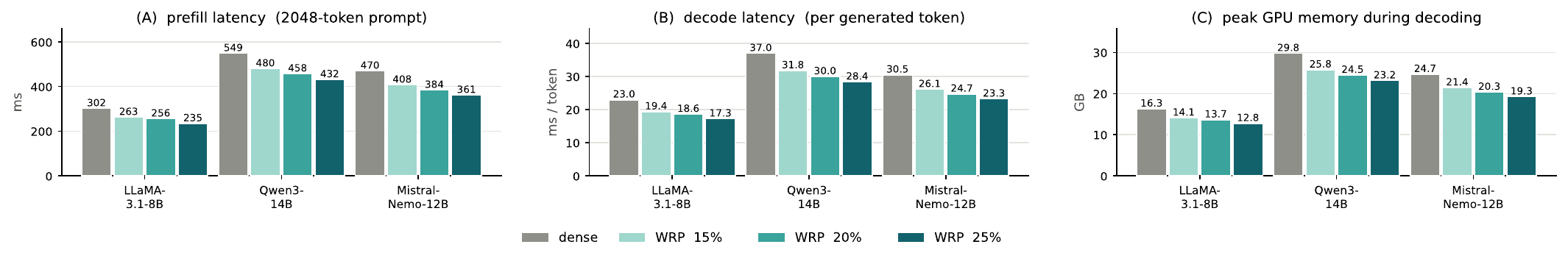}
    \caption{Inference cost of dense and WRP-pruned models. Percentages denote depth-removal budgets; panels show prefill latency, per-token decode latency, and peak allocated GPU memory.}
    \label{fig:latency}
\end{figure*}

%% file: contents/4-results-full-table.tex
\begin{table*}[t]
\centering
\small
\begin{adjustbox}{width=1\textwidth}
\begin{tabular}{l l l c c c c c c c c c c r}
\toprule
\textbf{Model} & \textbf{$L_p/L_t$} & \textbf{Method} & \textbf{Forward} & \textbf{ARC-E} & \textbf{ARC-C} & \textbf{HellaS} & \textbf{WinoG} & \textbf{BoolQ} & \textbf{OBQA} & \textbf{RTE} & \textbf{COPA} & \textbf{RACE} & \textbf{Avg.}$\uparrow$ \\
\midrule
\multirow{11}{*}{\begin{sideways}LLaMA-3.1-8B\end{sideways}}
& 0/32 & Dense & O & 81.14 & 53.50 & 78.89 & 73.56 & 82.08 & 44.80 & 69.31 & 87.00 & 39.14 & 67.71 \\
\cmidrule{2-14}
& 6/32 & LLM-Streamline & O & 64.56 & 44.71 & 67.50 & 68.19 & 70.06 & 40.20 & 58.12 & 81.00 & 36.27 & 58.96 \\
& 6/32 & ShortGPT & O & 62.33 & 43.77 & 68.22 & 68.51 & 71.99 & 37.40 & 66.43 & 79.00 & 35.02 & 59.19 \\
& 6/32 & LoRP & O & 69.49 & 42.75 & 67.56 & 68.90 & 66.12 & 37.00 & 66.06 & 86.00 & 37.42 & \textbf{60.14} \\
& 6/32 & Mag+ & X & 51.68 & 27.99 & 45.96 & 51.62 & 51.44 & 32.60 & 57.04 & 69.00 & 27.75 & 46.12 \\
& 6/32 & \wrpshade WRP (Ours) & \wrpshade X & \wrpshade 67.89 & \wrpshade 45.31 & \wrpshade 66.89 & \wrpshade 67.56 & \wrpshade 67.34 & \wrpshade 38.20 & \wrpshade 59.57 & \wrpshade 83.00 & \wrpshade 36.08 & \wrpshade 59.09 \\
\cmidrule{2-14}
& 8/32 & LLM-Streamline & O & 41.54 & 31.83 & 30.89 & 54.22 & 37.61 & 29.00 & 64.62 & 64.00 & 25.26 & 42.11 \\
& 8/32 & ShortGPT & O & 41.54 & 31.83 & 30.89 & 54.22 & 37.61 & 29.00 & 64.62 & 64.00 & 25.26 & 42.11 \\
& 8/32 & LoRP & O & 59.01 & 40.19 & 58.68 & 63.14 & 69.94 & 34.20 & 55.60 & 76.00 & 32.25 & \textbf{54.33} \\
& 8/32 & Mag+ & X & 48.53 & 25.77 & 41.74 & 54.22 & 44.53 & 29.60 & 47.65 & 70.00 & 26.22 & 43.14 \\
& 8/32 & \wrpshade WRP (Ours) & \wrpshade X & \wrpshade 59.55 & \wrpshade 39.59 & \wrpshade 60.65 & \wrpshade 62.75 & \wrpshade 65.57 & \wrpshade 35.80 & \wrpshade 53.07 & \wrpshade 73.00 & \wrpshade 34.45 & \wrpshade 53.82 \\
\midrule
\multirow{11}{*}{\begin{sideways}Qwen3-14B\end{sideways}}
& 0/40 & Dense & O & 82.79 & 60.24 & 78.81 & 73.17 & 89.33 & 46.20 & 77.62 & 90.00 & 43.25 & 71.27 \\
\cmidrule{2-14}
& 8/40 & LLM-Streamline & O & 67.42 & 40.27 & 62.02 & 58.33 & 70.00 & 40.80 & 62.82 & 71.00 & 33.97 & 56.29 \\
& 8/40 & ShortGPT & O & 64.90 & 41.81 & 57.07 & 62.27 & 80.46 & 35.20 & 64.98 & 70.00 & 40.29 & 57.44 \\
& 8/40 & LoRP & O & 68.56 & 45.90 & 62.35 & 65.51 & 75.81 & 34.20 & 62.82 & 75.00 & 37.32 & 58.61 \\
& 8/40 & Mag+ & X & 39.98 & 27.56 & 47.59 & 50.12 & 61.59 & 26.00 & 58.12 & 60.00 & 33.97 & 44.99 \\
& 8/40 & \wrpshade WRP (Ours) & \wrpshade X & \wrpshade 68.98 & \wrpshade 43.00 & \wrpshade 63.81 & \wrpshade 63.69 & \wrpshade 65.05 & \wrpshade 38.00 & \wrpshade 70.40 & \wrpshade 80.00 & \wrpshade 38.47 & \wrpshade \textbf{59.04} \\
\cmidrule{2-14}
& 10/40 & LLM-Streamline & O & 37.88 & 32.17 & 41.75 & 52.33 & 67.98 & 31.00 & 55.23 & 55.00 & 30.14 & 44.83 \\
& 10/40 & ShortGPT & O & 59.34 & 37.29 & 50.70 & 57.70 & 67.09 & 31.60 & 52.71 & 60.00 & 37.70 & 50.46 \\
& 10/40 & LoRP & O & 66.33 & 41.89 & 58.40 & 62.75 & 71.44 & 34.00 & 75.45 & 71.00 & 37.03 & \textbf{57.59} \\
& 10/40 & Mag+ & X & 30.72 & 23.21 & 29.55 & 51.30 & 55.54 & 26.20 & 48.38 & 60.00 & 24.50 & 38.82 \\
& 10/40 & \wrpshade WRP (Ours) & \wrpshade X & \wrpshade 61.83 & \wrpshade 38.40 & \wrpshade 57.64 & \wrpshade 62.43 & \wrpshade 66.61 & \wrpshade 34.20 & \wrpshade 69.31 & \wrpshade 78.00 & \wrpshade 35.31 & \wrpshade 55.97 \\
\midrule
\multirow{11}{*}{\begin{sideways}Mistral-Nemo-12B\end{sideways}}
& 0/40 & Dense & O & 81.65 & 57.94 & 82.79 & 73.40 & 85.32 & 47.20 & 64.98 & 91.00 & 41.91 & 69.58 \\
\cmidrule{2-14}
& 8/40 & LLM-Streamline & O & 65.87 & 43.69 & 70.78 & 72.45 & 66.48 & 36.80 & 53.43 & 84.00 & 38.09 & 59.07 \\
& 8/40 & ShortGPT & O & 67.26 & 45.22 & 71.28 & 72.38 & 65.78 & 38.00 & 59.93 & 86.00 & 39.33 & \textbf{60.57} \\
& 8/40 & LoRP & O & 65.70 & 43.09 & 71.24 & 68.82 & 67.52 & 38.40 & 57.40 & 88.00 & 37.99 & 59.80 \\
& 8/40 & Mag+ & X & 41.79 & 24.74 & 41.19 & 52.80 & 54.40 & 29.40 & 53.43 & 71.00 & 23.83 & 43.62 \\
& 8/40 & \wrpshade WRP (Ours) & \wrpshade X & \wrpshade 68.35 & \wrpshade 44.28 & \wrpshade 69.87 & \wrpshade 69.14 & \wrpshade 64.10 & \wrpshade 38.00 & \wrpshade 61.37 & \wrpshade 86.00 & \wrpshade 38.09 & \wrpshade 59.91 \\
\cmidrule{2-14}
& 10/40 & LLM-Streamline & O & 61.32 & 41.38 & 63.84 & 71.43 & 66.67 & 35.20 & 58.48 & 78.00 & 38.66 & \textbf{57.22} \\
& 10/40 & ShortGPT & O & 63.30 & 44.20 & 58.72 & 63.06 & 67.22 & 35.40 & 63.54 & 79.00 & 34.83 & 56.59 \\
& 10/40 & LoRP & O & 61.49 & 40.27 & 64.43 & 64.40 & 66.54 & 37.40 & 63.90 & 80.00 & 35.41 & 57.09 \\
& 10/40 & Mag+ & X & 39.86 & 24.91 & 33.69 & 50.91 & 55.05 & 27.00 & 50.18 & 70.00 & 22.01 & 41.51 \\
& 10/40 & \wrpshade WRP (Ours) & \wrpshade X & \wrpshade 64.44 & \wrpshade 40.02 & \wrpshade 64.29 & \wrpshade 65.35 & \wrpshade 65.60 & \wrpshade 35.40 & \wrpshade 55.60 & \wrpshade 84.00 & \wrpshade 34.93 & \wrpshade 56.62 \\
\bottomrule
\end{tabular}
\end{adjustbox}
\caption{Accuracy (\%) on all nine tasks after one-shot, recovery-free depth pruning. Forward marks calibration forward passes; bold marks the best pruned average per setting. $L_p / L_t$ denotes the number of pruned blocks over the total number of blocks.}
\label{tab:main}
\end{table*}

%% file: contents/3-method.tex
\label{sec:method}

\noindent\textbf{Preliminaries.}
Consider a pretrained LLM with $L$ Transformer blocks indexed by $\ell\in\{0,\ldots,L-1\}$. Let $P$ denote the number of blocks to remove and $\mathcal D$ the removal set, with $|\mathcal D|=P$. Let $W_\ell^a$ denote the weight matrix of projection $a$ in block $\ell$. For $a\in\{o,\mathrm{down}\}$, $W_\ell^a\in\mathbb R^{d\times m_a}$, where $d$ is the shared hidden-state dimension and $m_a$ is the input dimension of projection $a$. We center these matrices across output channels using $H_d=I_d-\frac{1}{d}\mathbf1\mathbf1^\top$, the centering step in the definition of linear CKA~\cite{kornblith2019cka}. WRP constructs a pairwise layer-similarity matrix $S\in\mathbb R^{L\times L}$ to determine $K$ layer clusters and select $\mathcal D$. We set $\epsilon=10^{-12}$.

\subsection{Weight-Space Block Descriptors}
\noindent\textbf{Cross-layer projection similarity.}
A Transformer block updates the hidden state through its attention output and MLP down projections,
$h_{\ell+1}=h_\ell+W_\ell^{o}(\cdot)+W_\ell^{\mathrm{down}}(\cdot)$.
Since pruning removes these contributions, we compute cross-layer similarity from $W_\ell^{o}$ and $W_\ell^{\mathrm{down}}$, which share the same hidden-state output space across layers. Other projections map into layer-specific internal spaces and depend on the input distribution, so we use only their scale information in $S^{\mathrm{scale}}$. For $a\in\{o,\mathrm{down}\}$, we define
\begin{equation}
\widetilde W_\ell^a=H_dW_\ell^a,
\quad
G_\ell^a=\widetilde W_\ell^a(\widetilde W_\ell^a)^\top
\label{eq:gram}
\end{equation}
We compare the resulting output-space inner-product matrices using linear CKA~\cite{kornblith2019cka}:
\begin{equation}
s_{ij}^{a}=\frac{\langle G_i^a,G_j^a\rangle_F}
{\|G_i^a\|_F\|G_j^a\|_F+\epsilon},
\quad
S_{ij}^{\mathrm{proj}}=\frac{s_{ij}^{o}+s_{ij}^{\mathrm{down}}}{2}
\label{eq:weight_cka}
\end{equation}
The Gram construction is invariant to permutations of intermediate dimensions
since $(W\Pi)(W\Pi)^\top=WW^\top$, and CKA additionally normalizes differences
in scale.

\noindent\textbf{Projection-scale similarity.}
To capture relative attention and MLP scales, we represent each block using the Frobenius norms of its seven projections $a\in\{q,k,v,o,\mathrm{gate},\mathrm{up},\mathrm{down}\}$:
\begin{equation}
f_\ell=(\|W_\ell^a\|_F)_a,
\quad
\widetilde f_\ell=f_\ell-\frac{1}{L}\sum_{m=0}^{L-1}f_m
\label{eq:fingerprint}
\end{equation}
Centering removes the model-wide average scale. We then define
\begin{equation}
S_{ij}^{\mathrm{scale}}=
\frac{\widetilde f_i^\top\widetilde f_j}
{\|\widetilde f_i\|_2\|\widetilde f_j\|_2+\epsilon}, \text{ }
S_{ij}=\tfrac12(S_{ij}^{\mathrm{proj}}+S_{ij}^{\mathrm{scale}}), \text{ }
S_{ii}=1 \notag
\end{equation}
The resulting matrix $S$ captures pairwise layer relations using both projection similarity and relative scale information.

\subsection{Weight-Only Clustering and Allocation}
Following LoRP~\cite{yun2026lorp}, we cluster layers before allocating removals. We use $K\in\{2,4\}$ as coarse and fine clustering candidates and choose between them directly from $S$. Set $A=(S+1)/2$, with unit diagonal, and construct
\begin{equation}
\mathcal L=I-\Delta^{-1/2}A\Delta^{-1/2},
\quad
\Delta=\operatorname{diag}(A\mathbf1)
\label{eq:laplacian}
\end{equation}
Let $0=\lambda_1\leq\lambda_2\leq\cdots\leq\lambda_L$ be the ascending eigenvalues of $\mathcal L$ and define $d_j=\lambda_{j+1}-\lambda_j$ for $j\in\{2,3\}$. Here, $d_2$ measures the strength of a coarse two-cluster structure, while $d_3$ captures evidence for additional structure beyond it. To reduce depth-proximity effects, we construct $S_0$ by replacing each $S_{ij}$ with the mean similarity at the same depth distance $|i-j|$, while keeping a unit diagonal. Applying the same construction to $S_0$ gives reference gaps $n_2$ and $n_3$. Let $e_j=\max(d_j-n_j,0)$. When $e_2+e_3>0$, we compute
\begin{equation}
\rho=\frac{e_2}{e_2+e_3},
\quad
K=\begin{cases}
2,&\rho\geq\tfrac12\\
4,&\rho<\tfrac12
\end{cases}
\label{eq:binary_k}
\end{equation}
When $e_2+e_3=0$, we set $\rho=1$ and use $K=2$. Thus, when the coarse two-cluster structure dominates, we use $K=2$. Otherwise, we use the finer $K=4$ candidate.

Spectral clustering of $A$ produces $K$ clusters $\mathcal C_k$ using the normalized-Laplacian embedding and $k$-means with a fixed seed. Clusters are ordered by the depth of their shallowest block. We adopt LoRP's two-stage allocation rule~\cite{yun2026lorp}. For each block, we measure its average similarity to the other blocks in the same cluster. We also measure the average pairwise similarity among the remaining blocks $\mathcal R_k=\mathcal C_k\setminus\mathcal D$ in each cluster:
\begin{equation}
r(\ell;\mathcal C_k)=\frac{\sum_{m\in\mathcal C_k\setminus\{\ell\}}S_{\ell m}}
{|\mathcal C_k|-1},
\quad
\bar r(\mathcal R_k)=\frac{2\sum_{i<j,\,i,j\in\mathcal R_k}S_{ij}}
{|\mathcal R_k|(|\mathcal R_k|-1)}
\notag
\end{equation}
We protect the first and last blocks. Eligible blocks in each cluster follow a fixed ranking from highest to lowest $r$. Stage~1 selects the highest-ranked eligible block from each cluster in cluster order. Stage~2 repeatedly selects the highest-ranked eligible block from the cluster with the largest $\bar r$, updating $\mathcal R_k$ and $\bar r$ after each selection. Both stages skip clusters with no eligible blocks left and stop when $|\mathcal D|=P$. We set $\bar r=-\infty$ for $|\mathcal R_k|<2$ and break cluster ties in favor of the lower index. Selected blocks are removed without changing the remaining parameters.

%% file: contents/4-results.tex
\label{sec:experiments}


\subsection{Experimental Settings}
We evaluate LLaMA-3.1-8B (32 blocks)~\cite{grattafiori2024llama3}, Qwen3-14B (40)~\cite{yang2025qwen3}, and Mistral-Nemo-12B (40)~\cite{mistral2024nemo}, with pruning budgets of 6 and 8 blocks for LLaMA, and 8 and 10 blocks for Qwen3 and Mistral-Nemo. We report zero-shot accuracy on nine benchmarks using LM Evaluation Harness: ARC-Easy, ARC-Challenge~\cite{clark2018arc}, HellaSwag~\cite{zellers2019hellaswag}, WinoGrande~\cite{sakaguchi2021winogrande}, BoolQ~\cite{clark2019boolq}, OpenbookQA~\cite{mihaylov2018openbookqa}, RTE~\cite{rte}, COPA~\cite{roemmele2011copa}, and RACE~\cite{lai2017race}. Equation~\eqref{eq:binary_k} gives $(\rho,K)=(0.778,2)$, $(0.084,4)$, and $(1.000,2)$, respectively. All methods are recovery-free. We compare against activation-based LLM-Streamline, ShortGPT, and LoRP, and forward-free Mag+. Mag+ follows the official Shortened LLaMA setting~\cite{kim2024shortened}, protecting the first four and last two blocks, while WRP protects only the first and last.

\subsection{Main Results}
At matched budgets, WRP outperforms Mag+ across all six settings by 10.68--17.15 points, with an average gain of 14.38 points (Table~\ref{tab:main}). WRP averages 57.41 compared with 57.93 for LoRP and achieves the best pruned average on Qwen3-14B at 8/40. These results show that pairwise weight relations enable strong forward-free layer selection while remaining close to activation-based methods. Figure~\ref{fig:patterns} shows the resulting non-contiguous removal patterns across depth.

\subsection{Inference Cost}
\label{sec:latency}
We measure inference cost on a single NVIDIA RTX~6000 Ada Generation GPU using FP16 and batch size~1. At 25\% pruning, WRP reduces prefill latency by 21--23\%, decode latency by 23--25\%, and peak GPU memory by 21--22\% (Fig.~\ref{fig:latency}). These gains come directly from reducing model depth and require no custom kernels or runtime changes.

%% file: contents/6-conclusion.tex
WRP enables forward-free depth pruning by estimating inter-layer redundancy directly from checkpoint weights. Across six pruning settings, it consistently outperforms magnitude-based pruning while remaining close to activation-based methods. These results show that checkpoint weights alone provide useful signals for effective depth pruning.